# A Passivity Based Framework for Safe Physical Human Robot Interaction

Z. Ding, M. Baghbahari, and A. Behal

*Abstract*— In this paper, the problem of making a safe compliant contact between a human and an assistive robot is considered. Users with disabilities have a need to utilize their assistive robots for physical human-robot interaction (PHRI) during certain activities of daily living (ADLs). Specifically, we propose a hybrid force/velocity/attitude control for a PHRI system based on measurements from a 6-axis force/torque sensor mounted on the robot wrist. While automatically aligning the end-effector surface with the unknown environmental (human) surface, a desired commanded force is applied in the normal direction while following desired velocity commands in the tangential directions. A Lyapunov based stability analysis is provided to prove both convergence as well as passivity of the interaction to ensure both performance and safety. Simulation as well as experimental results verify the performance and robustness of the proposed hybrid controller in the presence of dynamic uncertainties as well as safe physical human-robot interactions for a kinematically redundant robotic manipulator.

## I. INTRODUCTION

While assistive robotic devices such as Wheelchair Mounted Robotic Arms (WMRAs) [1]-[5] and Companion Robots [6]-[10] traditionally help users with object retrieval [11] or pick and place tasks [12], they are quite capable of physical human-robot interaction (PHRI) with the user themselves. Users with disabilities have a need for assistance with activities of daily livings (ADLs) such hair-grooming, scratching, face-sponging *etc.*; all these daily activities require physical interaction with various surfaces on the human body. Under this need, the assistive robot has to be able to align with the unknown (human) surface, and also apply a desired force in the normal direction while following the surface based on desired velocity profiles that the user can command to the robot. It is critical that the assistive robot be able to execute a safe compliant contact with the human user.

Researchers have proposed various methods to achieve hybrid position/force control on a surface. In [13], the author used the exact CAD model for polishing position/force control. With the CAD/CAM model, they are able to track the desired trajectory, force, and contact direction. Since it requires an exact CAD model of the environment, this method will not work for the PHRI problem considered in this paper. In [14], the authors designed and implemented a compliant arm to perform bed bath for patient hygiene. A bang-bang controller

Z. Ding and M. Baghbahari are graduate students with the Electrical and Computer Engineering Dept. at the University of Central Florida (UCF), Orlando, FL 32816. (E-mail: zhangchi@knights.ucf.edu; baghbahari@gmail.com)

Aman Behal is with ECE and NanoScience Technology Center at UCF, Orlando, FL. Email: abehal@ucf.edu (corresponding author)

was utilized to maintain the z-axis force against the body between 1-3N while a laser range finder was utilized to retrieve the skin surface point cloud of the skin followed by selection of wiping area by the operator. This method also required to obtain the point cloud of the environment. The authors of [15] proposed a contact force model for wiping and shaving tasks. They captured the face point cloud and the force profile of health participants performing daily living tasks such as wiping and shaving. Then, they built a three-parameter trapezoidal force model of each stroke and the force dependency on the facial area. Besides assuming previous knowledge of the environment, there are other approaches for unknown environments. In [16], the author proposed two methods for exploring unknown surfaces with discontinuities by using only a force/torque sensor. They rotate the direction of the desired motion/force instead of rotating the end-effector to keep moving and inserting force on unknown surface; however, this method cannot be used for certain PHRIs which need continuously variable alignment between the end-tool and human body, such as during shaving. In [17], the authors proposed a hybrid position-force sliding mode control for surface treatment such as polishing, grinding, finishing, and deburring – the end-effector can apply the desired pressure on the surface and also keep the end-effector orientation perpendicular to the surface; however, the orientation constraint is not considered for moving along a frictional environment. In [18], the authors proposed a deformation-tracking impedance control for interacting with unknown surfaces by using an extended Kalman filter to estimate the parameters of the environment, thereby controlling the interaction force indirectly by tracking the desired deformation without force sensing. However, this method cannot estimate the interaction torque; therefore, during the alignment phase of the assembly task, their desired interaction torque is just determined experimentally. In [19], the authors proposed an inverse differential kinematics-based position/force control for cleaning an unknown surface. They utilized a force/torque sensor to provide feedback for the force control part. However, this velocity control-based algorithm is not considered safe for human-robot interaction; furthermore, the evaluation of surface alignment is also missing.

In this paper, we propose a robust hybrid force/velocity/attitude controller which guarantees passivity of a robot manipulator with respect to unknown environments. By using measurements of robot states as well as interaction forces/torques at the end-effector, we are able to reshape the end-effector impedance (namely inertia, damping, and stiffness) to desired values needed for a safe and compliant interaction. Sliding mode control is utilized to compensate for exogenous disturbances, model uncertainty, and friction effects which can otherwise degrade the performance of the impedance controller. Within the sliding mode controller, we are able to achieve the desired system dynamics with a general assistive robot which are lower cost and less precision than industry collaboration robot. The resulting controller is able to align the end-effector with the unknown environmental surface while simultaneously tracking desired force and velocity profiles, respectively, in the normal and tangential directions. This work is novel in terms of the guarantees of 6-DOF exponential stability and passivity of our hybrid controller while interacting with an unknown environment.

The remainder of this paper is paper is organized as follows. Sections II and III deal, respectively, with the problem statement and the modeling. In Section IV, we present the controller design, stability analysis, proof of passivity and simulation results for a frictionless environment. In Section V, we present the corresponding control design and stability analysis for a real environment with friction. Experimental results with a kinematically redundant collaborative robot are presented in Section VI. Section VII concludes the paper.

## II. Problem Statement

The research objective is to align the robot end-effector with the unknown environment and apply a desired force in the normal direction while following a commanded velocity profile along the tangential directions. In order to guarantee safe human-robot interaction, another research objective is to ensure that the robot acts as a passive system while transmitting user intent to and during interaction with the environment. To design and implement our robust impedance control framework, we assume knowledge of the joint position/velocity measurements as well as the interaction force at the end-effector using a wrist mounted 6-axis force/torque sensor. We assume uncertainty in the robot dynamics and no prior knowledge of the location/orientation of the environmental surface with respect to the robot coordinate system. While we assume that the surface presents damping and stiffness in the normal direction and pure damping along the surface, we assume no prior knowledge of the parameters.

## III. Modeling

### A. Manipulator Model

The dynamics of an $n$ degree-of-freedom robot are given by

$$M(q)\ddot{q} + C(q,\dot{q})\dot{q} + G(q) = \tau + \tau_{env} - \tau_f \quad (1)$$

where $M(q) \in \mathbb{R}^{n \times n}$ is the symmetric positive definite inertia matrix, $C(q,\dot{q}) \in \mathbb{R}^{n \times n}$ is the matrix of Coriolis and centrifugal torques, $G(q) \in \mathbb{R}^n$ is the vector of gravitational torques, $q, \dot{q}, \ddot{q} \in \mathbb{R}^n$ denote, respectively, the joint angle, joint velocity and joint acceleration vectors, $\tau \in \mathbb{R}^6$ is the control input vector of joint torques, $\tau_{env} = J^T F_{e,e} \in \mathbb{R}^6$ is the external torque registered at the robot joints, $F_{e,e} \triangleq \begin{bmatrix} f_{e,e}^T & \tau_{e,e}^T \end{bmatrix}^T = \begin{bmatrix} f_{e,x} & f_{e,y} & f_{e,z} & \tau_{e,x} & \tau_{e,y} & \tau_{e,z} \end{bmatrix}^T \in \mathbb{R}^6$ is the interaction force measured by the force/torque sensor mounted on the wrist, $J \in \mathbb{R}^{6 \times n}$ is the Jacobian matrix, while $\tau_f \in \mathbb{R}^6$ denotes joint friction. The joint velocity and acceleration for a redundant robot (*i.e.*, $n > 6$) can be written as follows

$$\dot{q} = J^+\dot{x} + (I - J^+J)b \quad (2)$$
$$\ddot{q} = J^+\ddot{x} - J^+\dot{J}J^+\dot{x} - J^+\dot{J}(I - J^+J)b \quad (3)$$

where $\dot{x} = \begin{bmatrix} v_b^T & \omega_b^T \end{bmatrix}^T$, $\ddot{x} = \begin{bmatrix} \dot{v}_b^T & \dot{\omega}_b^T \end{bmatrix}^T \in \mathbb{R}^6$ denote end-effector velocity and acceleration vectors, respectively, $v_b$ and $\omega_b$ are the end-effector translation and angular velocity expressed in the base frame, $J^+ \triangleq J^T(JJ^T)^{-1}$ denotes the right pseudoinverse of the Jacobian matrix, while $b \in \mathbb{R}^n$ is an arbitrary vector utilized to accomplish secondary objectives such as joint limit, collision avoidance, *etc.* For ease of presentation, we choose $b = 0$ for the remainder of the paper. After replacing the joint acceleration and velocity by (2) and (3), we can obtain the task space robot dynamics as follows

$$MJ^+\ddot{x} - MJ^+\dot{J}J^+\dot{x} + CJ^+\dot{x} + G - J^T F_{e,e} + \tau_f = \tau \quad (4)$$

To accomplish our velocity tracking objective, we define error $e \triangleq \begin{bmatrix} e_v & e_\omega \end{bmatrix}^T \in \mathbb{R}^6$ in the end-effector frame as follows

$$e = \begin{bmatrix} v_e^T & \omega_e^T \end{bmatrix}^T - \begin{bmatrix} v_d^T & 0_{1 \times 3} \end{bmatrix}^T \quad (5)$$

where $v_d(t) \triangleq \begin{bmatrix} v_{d,x}(t) & v_{d,y}(t) & 0 \end{bmatrix}^T \in \mathbb{R}^3$ denotes the desired translational velocity in the end-effector frame, $v_e = \begin{bmatrix} v_{e,x} & v_{e,y} & v_{e,z} \end{bmatrix}^T = (R)^T v_b$ and $\omega_e \triangleq \begin{bmatrix} \omega_{e,x} & \omega_{e,y} & \omega_{e,z} \end{bmatrix}^T = (R)^T \omega_b$ are, respectively, the end-effector translation and angular velocity expressed in the end-effector frame, while $R(t) \in SO(3)$ denotes the rotation matrix between the robot base frame and the end-effector frame. By utilizing the velocity tracking error expressed in the end-effector frame, $e(t)$ can be related to the end-effector velocity variables in the base frame as follows

$$e = \underbrace{\begin{bmatrix} R & 0 \\ 0 & R \end{bmatrix}^T}_{\bar{R}} \underbrace{\begin{bmatrix} v_b \\ \omega_b \end{bmatrix}}_{\dot{x}} - \underbrace{\begin{bmatrix} v_d \\ 0_{3 \times 1} \end{bmatrix}}_{\bar{v}_d}.$$

By rearranging (5) and taking its time derivative, one can obtain the following expressions for $\dot{x}(t)$ and $\ddot{x}(t)$

$$\dot{x} = \bar{R}e + \bar{R}\bar{v}_d \quad (6)$$
$$\ddot{x} = \bar{R}\dot{e} + \bar{R}\dot{\bar{v}}_d + \dot{\bar{R}}e + \dot{\bar{R}}\bar{v}_d \quad (7)$$

By substituting (6) and (7) in (4), we can obtain the open-loop error dynamics as follows

$$MJ^+\bar{R}\dot{e} = \tau + \tau_{env} - \tau_f - G + U\dot{x} - MJ^+\bar{R}\dot{\bar{v}}_d \quad (8)$$

where $U \triangleq MJ^+\dot{J}J^+ - CJ^+ - MJ^+\dot{\bar{R}}\bar{R}^T$. We denote the best guess estimates of $M, G$, and $U$, respectively, as $\hat{M}, \hat{G}$, and $\hat{U}$, while $\tilde{M} \triangleq M - \hat{M}, \tilde{G} \triangleq G - \hat{G}$, and $\tilde{U} \triangleq U - \hat{U}$ denote the corresponding uncertainties. Based on these definitions, we can rewrite the open-loop error dynamics as follows

$$\hat{M}J^+\bar{R}\dot{e} = \tau + \tau_{env} - \hat{G} + \hat{U}\dot{x} - \hat{M}J^+\bar{R}\dot{\bar{v}}_d + \tilde{M}D \quad (9)$$

where

$$D \triangleq \hat{M}^{-1}(-\tilde{G} + \tilde{U}\dot{x} - \tilde{M}J^{+}\overline{\dot{R}v}_d - \tau_f) \qquad (10)$$

is a lumped model uncertainty term. Before proceeding further, we are motivated by the structure of the robot dynamics and the ensuing control development and stability analysis to assume the existence of the following properties:

**Property 1:** All kinematic singularities are always avoided and the pseudoinverse of the manipulator Jacobian, denoted by $J^{+}(q)$, is assumed to always exist.

**Property 2:** The actual and best estimate values of $M, G, U$ are assumed to be upper bounded as follows: $\|M\|_{i\infty} \leq b_M$, $\|G\|_{i\infty} \leq b_G$, $\|U\|_{i\infty} \leq b_{U_0} + b_{U_1}\|\dot{q}\|$, $\|\hat{M}\|_{i\infty} \leq b_{\hat{M}}$, $\|\hat{G}\|_{i\infty} \leq b_{\hat{G}}$, $\|\hat{U}\|_{i\infty} \leq b_{\hat{U}_0} + b_{\hat{U}_1}\|\dot{q}\|$ where $\|\cdot\|_{i\infty}$ denotes the induced infinity norm of a matrix while $b_M, b_G, b_{U_0}, b_{U_1}, b_{\hat{M}}, b_{\hat{G}}, b_{\hat{U}_0}, b_{\hat{U}_1}$ denote known positive bounding constants [22][23].

**Property 3:** The inverse of the inertia matrix is assumed to be bounded by a known positive constant $b_{M_{inv}}$ as $\|M^{-1}\|_{i\infty} \leq b_{M_{inv}}$ where $b_{M_{inv}}$ denotes a known positive bounding constant [23].

**Property 4:** Based on Properties 2 and 3, the lumped model uncertainty term $D$ defined above in (10) can be upper bounded by a function of the joint velocity as follows

$$\|D\| < b_{D0} + b_{D1}\|\dot{q}\| + b_{D2}\|\dot{q}\|^{2} \qquad (11)$$

where $b_{D0}, b_{D1}$, and $b_{D2}$ denote known positive bounding constants.

### B. Environment Model

We model the environment as a spring-damper which provides the environmental force in object (environment) frame as follows

$$f_{e,o} = K_e\hat{z}_v(z_n - z_o) + B_e v_o \qquad (12)$$

where $K_e = diag\begin{bmatrix} 0 & 0 & k_e \end{bmatrix}$, $B_e = diag\begin{bmatrix} b_e & b_e & 0 \end{bmatrix} \in \mathbb{R}^{3\times3}$ are diagonal matrices of environment stiffness and damping, respectively, $\hat{z}_v = \begin{bmatrix} 0 & 0 & 1 \end{bmatrix}^{T}$ denotes the standard basis vector in the z-direction, while $z_n \in \mathbb{R}$ is the z-axis neutral position of the environment in the object frame and $z_o \in \mathbb{R}$ is the current z-axis position of the end-effector expressed in the object frame. Here, $v_o = R_o^o v_e$ is the velocity of the end-effector with respect to the object expressed in the object frame, $v_e$ is the end-effector translational velocity as defined earlier, while $R_o^e$ is the unknown rotation matrix between the object frame and end-effector frame. In the end-effector frame, the environmentally exerted torque on the end-effector can be defined as follows

$$\tau_e = r_e \times f_{e,e} \qquad (13)$$

where $r_e$ is the unknown position vector from the center of the sensor to the contact point while $f_{e,e} = R_o^e f_{e,o}$ is the environmental force expressed in the end-effector frame. The model of the interaction between the end-effector and the environment is shown in Figure 1. We also model the end-tool for the manipulator as a rigid partial sphere as specified in Figure 2. In the figure, $x_{ee}$ denotes the center of the robot wrist where the 6-axis force/torque sensor is mounted, $x_c$ denotes the center of the sphere, $r_R$ is the position vector from the sphere center of the end-tool to the contact point, while $r_{off}$ is the position vector from the sphere center to the robot wrist center and is parallel with the end-effector z-axis denoted by $\hat{z}_e$.

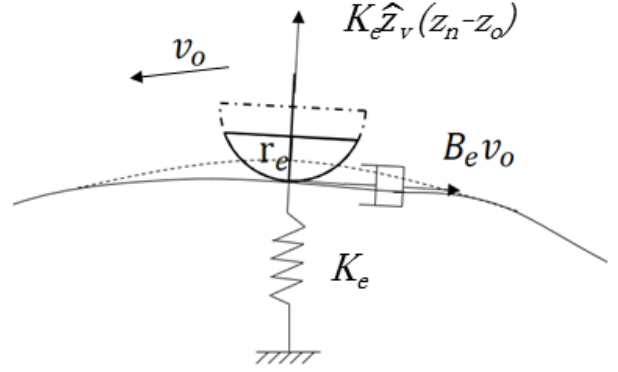

Fig. 1. Spring-damper environmental model

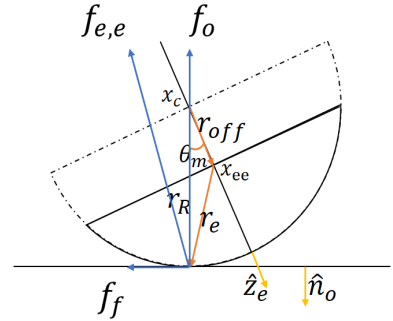

Fig. 2. End-tool geometry, where $f_o = K_e(x_n - x_o)$, $f_f = B_e v_o$ are the spring and damper induced forces from the environment.

## IV. Control Design and Stability Analysis: Non-dissipative Environment

### A. Control Design

The proposed control design has an inner- and an outer-loop. While the inner-loop is a robust controller to compensate for the system uncertainties and linearize the robot dynamics, the outer-loop reshapes the linearized dynamics to the desired dynamics. In what follows, we discuss the design of control strategies within the two loops that guarantees robust stability, convergence, as well as passivity. For lucidity of presentation, we first demonstrate PHRI stability and passivity when interacting with a non-dissipative (frictionless) environmental surface.

*1) Design of the inner loop:* Based on the structure of the open-loop error dynamics in (8) and our desire to obtain an impedance controller, we first design a computed torque inner loop controller to linearize the dynamics as follows

$$\tau = \hat{M}a_j - \tau_{env} + \hat{G} - \hat{U}\dot{x} + \hat{M}J^+\overline{R}\ddot{v}_d \qquad (14)$$

By substituting (14) into (9), we can obtain

$$J^+\overline{R}\dot{e} = a_j - D. \qquad (15)$$

In (14), $a_j$ is an auxiliary control term that is designed to compensate for the disturbance using a sliding mode controller as follows

$$a_j = J^+a_x - Qsign(S) \qquad (16)$$

where $a_x \triangleq [\ a_{x,v} \quad a_{x,\omega}\ ]^T$ is a yet to be designed auxiliary control term related to the desired dynamics. Motivated by the bound in (11) and in a manner similar to [22], we design the gain, $Q$, for the $sign()$ function (the standard signum) as follows

$$Q = b_{D0} + b_{D1}\|\dot{q}\| + b_{D2}\|\dot{q}\|^2 + \alpha, \qquad (17)$$

where $\alpha$ is a positive design constant, $b_{D0}, b_{D1},$ and $b_{D2}$ have been introduced earlier in (11), while $S$ denotes a sliding surface which is defined as follows

$$S \triangleq \dot{q} + \int_0^t J^+\left(\dot{J}J^+\dot{x} - \overline{R}\ddot{v}_d - \dot{\overline{R}}\dot{e} - \dot{\overline{R}}\ddot{v}_d - a_x\right)dt \qquad (18)$$

such that

$$\dot{S} = J^+(\overline{R}\dot{e} - a_x) \\ = -D - Qsign(S). \qquad (19)$$

Then, we have the following result for the inner loop:

*Lemma 1:* Given the robot system in (9) under the control law of (14) and (16), the sliding surface $S$ and its time derivative $\dot{S}$ will converge to zero in finite time $t_1$, and remain there for all subsequent time.

*Proof:* We define a positive-definite function $V_S$ as follows

$$V_S = \frac{1}{2}S^TS \qquad (20)$$

After time differentiating (20) along (19) and utilizing (11) and (17), we can obtain

$$\dot{V}_S = S^T(-D - Qsign(S)) \qquad (21) \\ \leq -\left(Q - \left(b_{D0} + b_{D1}\|\dot{q}\| + b_{D2}\|\dot{q}\|^2\right)\right)\|S\| \\ \leq -\alpha\|S\| \qquad (22)$$

Given that sliding mode control [21] has finite convergence time $t_1$, it is easy to conclude that $\lim_{t\to t_1} S = 0$, $\lim_{t\to t_1}\dot{S} = J^+\overline{R}\dot{e} - J^+a_x = 0 \Rightarrow \lim_{t\to t_1}\overline{R}\dot{e} - a_x = 0.$ ∎

*2) Design of the Desired Dynamics:* To meet our velocity and force control objectives while projecting desired impedance characteristics on the environment, we propose the following desired dynamics in the translational axes

$$M_d\dot{e}_v + B_de_v + K_d\int_0^t e_v(\tau)\,d\tau = e_f \qquad (23)$$

where $M_d \triangleq diag\{m_{d,xy}, m_{d,xy}, m_{d,z}\}$, $B_d \triangleq diag\{b_{d,xy}, b_{d,xy}, b_{d,z}\}$, $K_d \triangleq diag\{k_{d,xy}, k_{d,xy}, 0\}$ denote desired mass, damping, and stiffness matrices (note all nonzero matrix elements are positive design constants), while $e_f \triangleq [\ 0 \quad 0 \quad f_{e,z} - f_{d,z}\ ]$ denotes the force error with $f_{d,z}$ denoting the desired force in the z-direction. In order to achieve the desired dynamics of (23), the auxiliary control term $a_{x,v}$ can now be designed as follows

$$a_{x,v} = R\left\{M_d^{-1}\left[-B_de_v - K_d\int_0^t e_v(\tau)\,d\tau + e_f\right]\right\} \qquad (24)$$

For the angular axis, we design a quaternion based controller to align the end-effector with the unknown environment. The rotation between the end-effector z-axis $\hat{z}_e$ and the environmental normal $\hat{n}_o$ can be represented by a unit quaternion $\mathbf{q}(q, q_0)$ where $q_0 = \cos(\frac{\theta_m}{2}), q = \sin(\frac{\theta_m}{2})n$, $\theta_m$ and $n$ being the angle and the rotation axis between $\hat{z}_e$ and $\hat{n}_o$. The quaternion can be extracted from the torque and force feedback as follows. Based on the definition of the environment torque from (13) and the geometry of the end-tool in Figure (2), we can rewrite the environmental interaction torque as follows

$$\tau_e = r_e \times f_{e,e} = (R_o^er_R - r_{off}) \times f_{e,e}. \qquad (25)$$

Since the environmental surface is assumed frictionless and based on the geometry of the end tool shown in Figure 2, it is evident that $R_o^er_R \parallel f_{e,e} \parallel \hat{n}_o$ and $r_{off} \parallel \hat{z}_e$ where the notation $\parallel$ denotes parallel vectors. Then, we can simplify the relationship expressed in (25) as

$$\tau_e = f_{e,e} \times r_{off} = \|f_{e,e}\|\|r_{off}\|\hat{n}_o \times \hat{z}_e. \qquad (26)$$

Now, it is possible to calculate the rotation axis $n$ as follows

$$n = -\frac{\hat{n}_o \times \hat{z}_e}{\|\hat{n}_o \times \hat{z}_e\|} = -\frac{\tau_e}{\|\tau_e\|}. \qquad (27)$$

Since the rotation direction is already contained in the vector $n$, the range of the angle $\theta_m$ between $\hat{n}_o$ and $\hat{z}_e$ is $[0, \frac{\pi}{2})$. From (26), we can calculate the angle $\theta_m$ between $\hat{n}_o$ and $\hat{z}_e$ as

$$\theta_m = \arcsin\left(\frac{\|\tau_e\|}{\|f_{e,e}\|\|r_{off}\|}\right). \qquad (28)$$

The dynamics of the unit quaternion $\mathbf{q}(q, q_0)$ are as follows

$$\dot{q}_0 = -\frac{1}{2}\omega_e^Tq \\ \dot{q} = \frac{1}{2}(q_0\omega_e + q \times \omega_e) \qquad (29)$$

To reshape the angular dynamics in a desired mass/damper form, we begin by designing the desired rotational dynamics as follows

$$I_d\dot{\omega}_e + B_d\omega_e = u \qquad (30)$$

where $I_d \triangleq diag\{I_{d,x}, I_{d,y}, I_{d,z}\}$ is the desired rotational inertia, while $B_d$ and $u$ are, respectively, a damping matrix and an auxiliary control input which are yet to be designed. By taking the time derivative of a filtered tracking error $r \triangleq \omega_e + K_1 q$ ($K_1$ is diagonal positive definite) and premultiplying by $I_d$, we obtain

$$I_d \dot{r} = I_d \dot{\omega}_e + I_d K_1 \dot{q} \tag{31}$$

which, by substitution of (29) and (30), can be rewritten as

$$I_d \dot{r} = -B_d \omega_e + u + \frac{1}{2} I_d K_1 (q_0 \omega_e + q \times \omega_e). \tag{32}$$

Motivated by the ensuing stability analysis, we design $u$ as follows

$$u = B_d \omega_e - \frac{1}{2} I_d K_1 (q_0 \omega_e + q \times \omega_e) - \overline{P} r - q \tag{33}$$

such that

$$I_d \dot{r} = -\overline{P} r - q \tag{34}$$

where $\overline{P}$ is a positive definite matrix. Now, it is possible to rewrite the desired rotational dynamics in the following compact form

$$I_d \dot{\omega}_e + B_{d,\omega} \omega_e = \tau_a \tag{35}$$

where $B_{d,\omega} \triangleq \frac{1}{2} \left( q_0 I_d K_1 + 2\overline{P} + [q]_\times \right)$ is a positive time-varying damping matrix, while

$$\tau_a \triangleq -(\overline{P} K_1 + I) q = \frac{(\overline{P} K_1 + I)}{2 \|f_{e,e}\| \|r_{off}\| q_0} \tau_e$$

is an auxiliary torque vector which is well defined everywhere on the domain of the problem. Finally, we can design the auxiliary control term $a_{x,\omega}$ defined in (16) as follows

$$a_{x,\omega} = R I_d^{-1} [\tau_a - B_{d,\omega} \omega_e] \tag{36}$$

The overall proposed controller is shown in Figure 3.

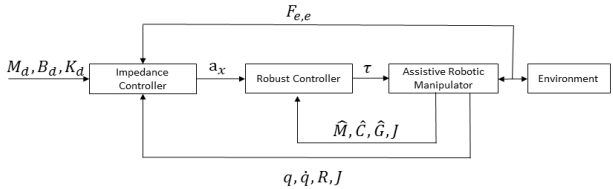

Fig. 3. Block Diagram of Proposed Physical Human Robot Interaction Controller

## B. Stability Analysis

Before presenting the main results, we state the following lemmas which will be invoked later.

*Lemma 2:* The desired dynamics of the angular axis in (35) are locally exponentially stable at the equilibrium point ($\omega_e = 0, q = 0, q_0 = 1$) in the sense that $\|q\| \leq \|q(0)\| e^{-\gamma t}$, $\|\omega_e\| \leq \|\omega_e(0)\| e^{-\gamma t}$ when $\|q(0)\| \in [0, 1/\sqrt{2})$ which implies that $\omega_e, q \in \mathcal{L}_1$ when the robot is interacting with a frictionless environment.

*Proof:* Inspired by [20], we define a nonnegative function $V_\omega$ as follows

$$V_\omega = q^T q + (q_0 - 1)^2 + r^T I_d r \tag{37}$$

which can be upperbounded as follows

$$V_\omega \leq \max\{2, \lambda_{\max}(I_d)\}(\|q\|^2 + \|r\|^2) \tag{38}$$

by utilizing the fact that $q^T q + q_0^2 = 1$. After taking the derivative of (37) along (34) and (29) and canceling common terms, we can obtain

$$\dot{V}_\omega = -q^T K_1 q - r^T \overline{P} r \tag{39}$$

It is easy to see from (37) and (39) that $q, r \in \mathcal{L}_2 \cap \mathcal{L}_\infty$. From this and previous definitions, it follows that $\omega_e, \dot{q}, \dot{r} \in \mathcal{L}_\infty$. Then, one can utilize Barbalat's Lemma [23] to prove that $\lim_{t \to \infty} q(t) = \lim_{t \to \infty} r(t) = 0$ which implies that $\lim_{t \to \infty} \omega_e = 0$, $\lim_{t \to \infty} q_0 = 1$. Since $\dot{V}_\omega$ can be lowerbounded as follows

$$\dot{V}_\omega \leq -\min\{\lambda_{\min}(K_1), \lambda_{\min}(\overline{P})\}(\|q\|^2 + \|r\|^2) \tag{40}$$

it is easy to see from (38) and (40) that $\dot{V}_\omega \leq -\gamma V_\omega$ where $\gamma = \frac{\min\{\lambda_{\min}(K_1), \lambda_{\min}(\overline{P})\}}{\max\{2, \frac{\lambda_{\max}(I_d)}{2}\}}$. Thus we can conclude that the equilibrium point ($q_0 = 1, q = 0, \omega_e = 0$) is exponentially stable in the sense specified in the statement of the Lemma. ∎

*Lemma 3:* The desired dynamics of the translational axes in (23) are globally exponentially stable (GES) at the equilibrium point ($e_v = 0, \dot{e}_v = 0$) when the robot is interacting with a frictionless environment. Furthermore, $v_{e,z}, \dot{v}_{e,z}, \dot{e}_{v,xy}, e_{v,xy} \in \mathcal{L}_1 \cap \mathcal{L}_\infty$ while $f_{e,o}, \tau_e \in \mathcal{L}_\infty$.

*Proof:* We begin our analysis in the x- and y-axes. The desired dynamics in the x- and y-axes are as follows

$$\begin{bmatrix} M_{d,xy} \dot{e}_{v,xy} \\ e_{v,xy} \end{bmatrix} = \begin{bmatrix} -B_{d,xy} & -K_{d,xy} \\ I_2 & 0_2 \end{bmatrix} \begin{bmatrix} e_{v,xy} \\ \int_0^t e_{v,xy}(\tau) \, d\tau \end{bmatrix} \tag{41}$$

where the notation is obvious from context given the previous definitions. Since (41) is a second-order LTI system with eigenvalues $\frac{-b_{d,xy} \pm \sqrt{b_{d,xy}^2 - 4k_{d,xy} m_{d,xy}}}{2m_{d,xy}}$ in the strict LHP, the equilibrium point ($e_{v,xy} = 0, \int_0^t e_{v,xy} d\tau = 0$) is GES. As for the z-axis, the desired dynamics of (23) in that direction can be written as follows

$$m_{d,z} \dot{v}_{e,z} = -b_{d,z} v_{e,z} + \hat{z}_v^T R_o^e f_{e,o} - f_{d,z}. \tag{42}$$

Since we are assuming no environmental friction, we can simplify (12) as

$$f_{e,o} = K_e \hat{z}_v (z_n - z_o) = \hat{z}_v k_e (z_n - z_o) \tag{43}$$

the time derivative of which can be written as

$$\dot{f}_{e,o} = -K_e \dot{z}_o = -K_e R_e^o v_e \tag{44}$$

To facilitate the analysis, we take the time derivative of (42) and obtain the dynamics as

$$m_{d,z} \ddot{v}_{e,z} = -b_{dz} \dot{v}_{e,z} - k_e v_{e,z} + G_1 \tag{45}$$

which is a second-order linear system with perturbation $G_1$ defined in (73) in the Appendix. By utilizing (74), (75), and (76), we can upperbound $G_1$ as follows

$$\|G_1\| \leq \max\left(\gamma_1, \gamma_2, \gamma_3\right) \exp\left(-\gamma t\right) \|x\| \qquad (46)$$

where $x \triangleq \begin{bmatrix} \dot{v}_{e,z} & v_{e,z} & 1 \end{bmatrix}^T$ is a state vector. Given this definition, we rewrite (45) as follows

$$\dot{x} = Ax + \hat{x}_v m_{d,z}^{-1} G_1 \qquad (47)$$

where $\hat{x}_v \triangleq \begin{bmatrix} 1 & 0 & 0 \end{bmatrix}^T$ denotes the standard basis vector in the x-direction while $A$ is a Hurwitz matrix defined in the Appendix. We also define a positive definite function $V = \frac{1}{2}x^T P x$ whose derivative along (47) is

$$\dot{V} = -x^T Q x + \frac{1}{2} x^T P \hat{x}_v m_{d,z}^{-1} G_1 + \frac{1}{2}(\hat{x}_v m_{d,z}^{-1} G_1)^T P x \qquad (48)$$

where $P$ and $Q$ are defined in the Appendix. After utilizing the growth bound of (46), we can upperbound (48) as follows

$$\dot{V} \leq -\left[\underbrace{\lambda_{\min}(Q)}_{\kappa_1} - \underbrace{\|P\| m_{d,z}^{-1} \max\left(\gamma_1, \gamma_2, \gamma_3\right)}_{\kappa_2} \exp\left(-\gamma t\right)\right] \|x\|^2$$

which implies exponential stability of $V(t)$ in the following sense

$$V(t) \leq V(0) \exp\left(\frac{-2\left(\kappa_1 t + \kappa_2 \gamma^{-1} e^{-\gamma t}\right)}{\lambda_{\min}(P)}\right) \left[u(t) - u(t - t_f)\right]$$
$$+ V(t_f) \exp\left(\frac{-2\left(\kappa_1(t-t_f) + \kappa_2 \gamma^{-1} e^{-\gamma\left(t-t_f\right)}\right)}{\lambda_{\max}(P)}\right) u(t - t_f) \qquad (49)$$

where $t_f = \max\left\{0, \frac{1}{\gamma} \ln\left(\frac{\kappa_2}{\kappa_1}\right)\right\}$ and $u(t)$ denotes the unit step function. Therefore, $x = 0$ is exponentially stable in the sense that $\|x\|, \|v_{e,z}\|, \|\dot{v}_{e,z}\| \leq \kappa e^{-\kappa_1 t}$ where $\kappa$ is a sufficiently large constant of analysis. Following the same process as in the proof for Lemma 2, it can be shown that $v_{e,z}, \dot{v}_{e,z}, \dot{e}_{v,xy}, e_{v,xy} \in \mathcal{L}_1 \cap \mathcal{L}_\infty$ which implies according to (42) that $f_{e,o}, f_{e,e} \in \mathcal{L}_\infty$ and that $\lim_{t \to \infty} f_{e,z} = f_{dz}$. Since $\tau_e = f_{e,e} \times r_{off} \leq \|f_{e,e}\| \|r_{off}\| \leq \sup\{\|f_{e,e}\|\} \|r_{off}\| < \infty$, it is evident that $\tau_e \in \mathcal{L}_\infty$. $\blacksquare$

*Remark 4:* From Lemma 1, it can be seen that the closed-loop dynamics converge to the desired dynamics in (23) and (35) such that the desired impedance characteristics can be projected on the frictionless environment. Based on Lemma 2, the quaternion $q$ (which represents the tool-environment misalignment) converges to zero which implies that the robot end-effector aligns with the frictionless environment. Furthermore, according to Lemma 3, the end-effector can reproduce the commanded end-effector velocities on the environmental tangential plane while applying the desired amount of force in a direction normal to the environment.

From the above Lemmas, we can also have the main passivity result for the proposed controller in the following theorem:

*Theorem 5:* The proposed control law can ensure the work done by the robot on the frictionless environment (human) is bounded in the sense that $W = \int_0^\infty (F_{env})^T v_e dt \leq c < \infty$ where $F_{env} \triangleq \begin{bmatrix} f_{e,e}^T & \tau_{e,e}^T \end{bmatrix}^T$ is the force/torque applied by the robot on the environment.

*Proof:* We split the work done by the robot into two phases: **(Part a)** Work done before reaching the desired dynamics: Since the sliding mode control has finite-time convergence, it will drive the system dynamics to the desired dynamics in finite time $t_1$. So the work done by the robot from $t = 0$ to $t = t_1$ is as follows

$$W_a = \int_0^{t_1} \begin{bmatrix} f_{e,e}^T & \tau_{e,e}^T \end{bmatrix} \begin{bmatrix} v_e^T & \omega_e^T \end{bmatrix}^T dt \qquad (50)$$

Based on the stability analysis of the sliding mode controller, we know that $S = \begin{bmatrix} S_v & S_\omega \end{bmatrix}, \dot{S} = \begin{bmatrix} \dot{S}_v & \dot{S}_\omega \end{bmatrix} \in \mathcal{L}_\infty$. During $[0, t_1]$

$$\dot{S} = J^+(\overline{R}\dot{e} - a_x) \neq 0.$$

Therefore, we can write

$$\dot{S}_v = J^+(R\dot{e}_v - a_{x,v}). \qquad (51)$$

Given the definition of (24), we can rewrite (51) as follows

$$M_d \dot{e}_v + B_d e_v + K_d \int_0^t e_v(\tau) d\tau - e_f = M_d R^T J \dot{S}_v$$

which is nothing but a perturbed version of the desired translational dynamics driven by the additional input $M_d R^T J \dot{S}_v$ which remains bounded between $[0, t_1]$ and is zero thereafter. Since we know from previous analysis that the solution for the desired dynamics of (23) stays bounded, the solution for the perturbed version driven by a bounded input over a finite time period will also stay bounded. A similar argument can be made for the angular velocity desired dynamics. Thus,

$$\begin{aligned} W_a &\leq \sup_t\{\|v_e\|\} \sup_t\{\|f_{e,e}\|\} t_1 + \\ &\quad \sup_t\{\|\omega_e\|\} \sup_t\{\|\tau_{e,e}\|\} t_1 \\ &\leq c_a \end{aligned} \qquad (52)$$

where $c_a$ is a positive constant of analysis. **(Part b)** After reaching the desired dynamics: The work done by the robot on the environment is denoted by $W_b(t)$, and can be bounded as follows

$$\begin{aligned} W_b &= \int_{t_1}^\infty (F_{env})^T v_e dt \\ &\leq \int_{t_1}^\infty \left\| \begin{bmatrix} f_{e,e}^T & \tau_{e,e}^T \end{bmatrix}^T \right\| \left\| \begin{bmatrix} v_e^T & \omega_e^T \end{bmatrix}^T \right\| dt \\ &\leq \sup_t\{\|v_{e,xy}\|\} \int_{t_1}^\infty \|f_{e,xy}\| dt + \sup_t\{\|f_{e,z}\|\} \int_{t_1}^\infty \|v_{e,z}\| dt \\ &\quad + \sup_t\{\|\tau_e\|\} \int_{t_1}^\infty \|\omega_e\| dt \end{aligned} \qquad (53)$$

since $v_{e,xy}, f_{e,z}, \tau_e \in \mathcal{L}_\infty$ as previously shown. Since $\|f_{e,xy}\| = \|A_{xy} R_o^e \hat{z}_v\| \|f_{e,o}\|$, we can bound

$$\begin{aligned} \int_{t_1}^\infty \|f_{e,xy}\| dt &\leq \sup_t\{\|f_{e,o}\|\} \int_{t_1}^\infty (\|2q_0 q_1 + 2q_2 q_3\| \\ &\quad + \|2q_1 q_3 - 2q_0 q_2\|) dt \end{aligned}$$

given that $f_{e,o} \in \mathcal{L}_\infty$ and where $A_{xy} \triangleq \begin{bmatrix} 1 & 0 & 0 \\ 0 & 1 & 0 \end{bmatrix}$. Since $\|q\| \leq 1$ and $q \in \mathcal{L}_1$, it is easy to see that $\int_{t_1}^\infty \|f_{e,xy}\| dt$ $< \infty$. Since $v_{e,z}, \omega_e \in \mathcal{L}_1$ as previously shown, it is clear to see that $W_b \le c_b < \infty$. Therefore, we can conclude that the total work done by the robot end-effector during Parts a and b is $W = W_a + W_b \le c_a + c_b \triangleq c < \infty$ which proves passivity. ∎

### C. Simulation Results

During the simulation, we choose 0-5 s as the initial alignment phase during which the commanded velocities in x- and y-directions are both chosen to be $0\,\mathrm{cm/s}$. Between 5-20 s and 20-35 s, we, respectively, command the desired velocity to be $\pm 1.5\,\mathrm{cm/s}$ along the end-effector y-direction. Finally, between 35-40 s, we command $0\,\mathrm{cm/s}$ velocity in the tangential direction of the end-effector. From Figures 4-8, we can see that the proposed controller can regulate the force error in end-effector z-direction to zero after the initial alignment process. The end-effector velocity along the tangential directions also tracks the desired velocity with nearly zero tracking error. As for the alignment, the norm of the quaternion $\|q\|$ and the misalignment angle decreases to 0. To evaluate the alignment, we define the so-called equivalent lever $r_{\mathrm{tan}} = \sqrt{F_z^2/(\tau_x^2 + \tau_y^2)}$ which represents the length of projection of the lever arm $r_e$ in end-effector x-y plane. When the end-effector is aligned with the environment, it is clear to see from Figure 8 that $r_{\mathrm{tan}}$ also converges to zero.

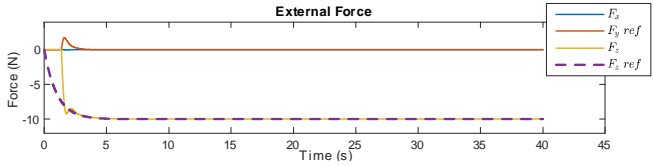

Fig. 4.   External force profile

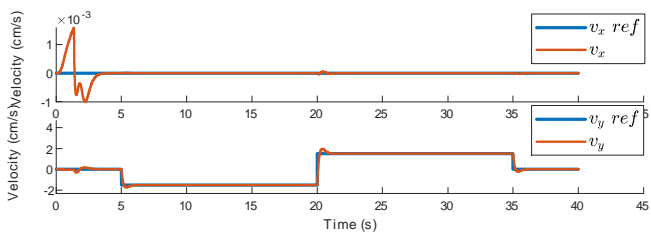

Fig. 5.   Velocity tracking profile

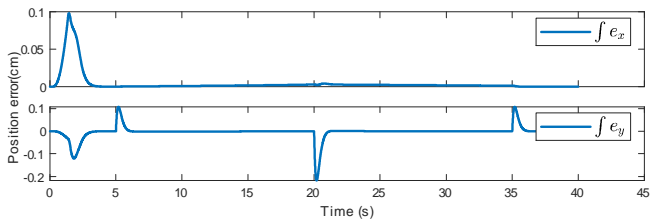

Fig. 6.   Position tracking profile

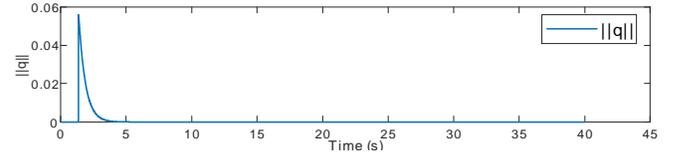

Fig. 7.   Quaternion tracking profile

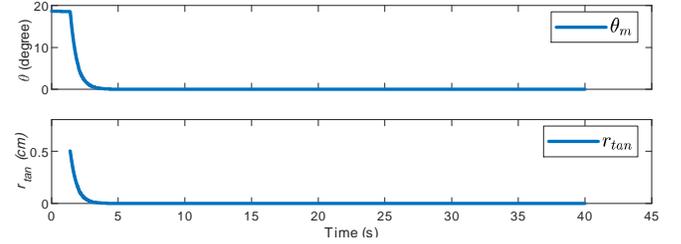

Fig. 8.   Misalignment evaluation. (Top) Misalignment angle between the end-effector z-direction and the normal of the contact surface, (Bottom) Equivalent lever in end-effector tangential direction.

## V. Control Design and Stability Analysis for Dissipative Environment

### A. Control Design

In the previous section, it is seen that the proposed controller is stable and passive during interaction with a frictionless environment. In this section, we will introduce the controller for a dissipative (frictional) environment which is a better model for the real world. For the case with friction, we can design the sliding mode control and torque input as similarly done in Section IV-A, but if friction is not compensated for, even though the equilibrium for the angular axis remains the same ($q_0 = 1, q = 0, \omega_e = 0$), the quaternion now represents the rotation between $r_e$ and $f_{e,e}$; therefore, the equilibrium represents $r_e \parallel f_{e,e}$. Since $f_{e,e} = f_o + f_f$ is the resultant of the friction force $f_f \in \mathbb{R}^3$ and the normal force $f_o \in \mathbb{R}^3$ from the object, the angle between the environment normal and $r_e$ is $\theta_o = \arctan(\frac{\|f_f\|}{\|f_o\|})$; per the geometry of the problem, there exists a non-zero misalignment angle $\theta_m$ between the end-effector and the environment normal which is related to the surface friction and the environment normal force. The orientation between the end-effector and environment can be seen in Figure 9.

$$\|r_{off}\| \sin(\theta_m) = \|r_e\| \sin(\theta_o)$$
$$\Rightarrow \theta_m = \arcsin\left(\frac{\|r_e\|}{\|r_{off}\|} \sin\left(\arctan\left(\frac{\|f_f\|}{\|f_o\|}\right)\right)\right) \quad (54)$$

To compensate for the misalignment due to friction, we design a desired torque $\tau_d \triangleq r_d \times f_{e,e}$ and extract the desired quaternion $\mathbf{q}_d(q_d, q_{d0})$ which represents the rotation between the desired lever arm $r_d \triangleq (\|r_R\| - \|r_{off}\|)\hat{z}_e$ and the environment force $f_{e,e}$. Similar to the case without friction in Section IV-A.2, we can define $\mathbf{q}(q, q_0)$ and $\mathbf{q}_d(q_d, q_{d0})$ as follows

$$q_0 = \cos(\frac{\theta_c}{2}), q = \sin(\frac{\theta_c}{2})n_e \quad (55)$$

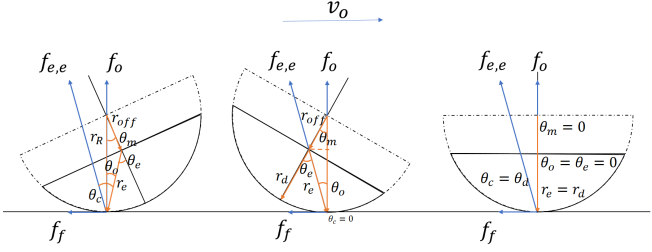

Fig. 9. Geometry of the end-tool. Left plot is the general misalignment case for the environment with friction. Middle plot is the equilibrium of the controller in last section applied to the frictional environment. The right plot is the desired equilibrium for the frictional environment.

$$q_{d0} = \cos(\frac{\theta_d}{2}), q_d = \sin(\frac{\theta_d}{2})n_d \qquad (56)$$

where $n_e = \frac{\tau_e}{\|\tau_e\|}, \sin(\theta_c) \triangleq \frac{\|\tau_e\|}{\|f_{e,e}\| \|r_e\|}$, $\theta_c$ is the angle between $r_e$ and $f_{e,e}$, while $n_d = \frac{\tau_d}{\|\tau_d\|}, \sin(\theta_d) \triangleq \frac{\|\tau_d\|}{\|r_d\| \|f_{e,e}\|}$, where $\theta_d$ is the angle between $r_d$ and $f_{e,e}$. Then, we define the quaternion error $\mathbf{e}(e, e_0)$ based on $\mathbf{q}$ and $\mathbf{q}_d$

$$\begin{aligned} \mathbf{e} &= \mathbf{q}_d^{-1} \circ \mathbf{q} \\ e_0 &= q_0 q_{d0} + q^T q_d \\ e &= q_{d0} q - q_0 q_d + [q]_\times q_d \end{aligned} \qquad (57)$$

The quaternion dynamics are as follows

$$\begin{aligned} \dot{e}_0 &= -\frac{1}{2}\omega_e^T e \\ \dot{e} &= \frac{1}{2}(e_0 \omega_e + e \times \omega_e) \end{aligned} \qquad (58)$$

With a process similar to that in Section IV-A.2, the desired dynamics of the rotation axis are designed as follows

$$I_d \dot{\omega}_e + B_{d,\omega f} \omega_e = \tau_{a,f} \qquad (59)$$

where

$$\begin{aligned} B_{d,\omega f} &= \frac{1}{2}(e_0 I_d K_1 + 2\overline{P} + [e]_\times) \\ \tau_{a,f} &= -(\overline{P}K_1 + 1)e \end{aligned}$$

such that the dynamics of $r \triangleq \omega_e + K_1 e$ can be written as

$$I_d \dot{r} = -\overline{P}r - e \qquad (60)$$

Finally, we design the auxiliary control term $a_{x,\omega}$ as follows

$$a_{x,\omega} = RI_d^{-1}[\tau_{a,f} - B_{d,\omega f}\omega_e] \qquad (61)$$

### B. Stability Analysis

*Lemma 6:* The desired dynamics of the angular axis in (59) is locally exponentially stable at the equilibrium point ($\omega_e = 0, e = 0, e_0 = 1$) when the robot is interacting with a frictional environment, in the sense that $\|e\| \le \|e(0)\| e^{-\gamma t}$, $\|\omega_e\| \le \|\omega_e(0)\| e^{-\gamma t}$ for $\|e(0)\| \in [0, 1/\sqrt{2})$ and a positive constant $\gamma$. Furthermore, the states of the angular axis $\omega_e, e \in \mathcal{L}_1$.

*Proof:* For the environment with friction, our sliding mode control remains the same, but the desired dynamics in angular axis are different. Based on the desired dynamics in

(59) and the dynamics of quaternion error in (58) we can define a positive function similar with (37) as follows

$$V_\omega = e^T e + (e_0 - 1)^2 + r^T I_d r \qquad (62)$$

where . After taking the time derivative of (62) along (58 ) and (60) and simplifying, we can have

$$\dot{V}_\omega = -e^T K_1 e - r^T \overline{P}r$$

With a process akin to that followed in the proof of Lemma 2 in Section IV-B, it is easy to see that the angular axis is exponentially stable at the equilibrium point ($e_0 = 1, e = 0, \omega_e = 0$) such that we can have $\|e\| \le \|e(0)\| e^{-\gamma t}$ and $\|\omega_e\| \le \|\omega_e(0)\| e^{-\gamma t}$ which clearly implies that $\omega_e, e \in \mathcal{L}_1$. ∎

*Lemma 7:* The desired dynamics of the translational axis in (23) are globally exponentially stable at the equilibrium point ($v_{e,e} = 0, \dot{v}_{e,e} = 0$) when the robot is interacting with a frictional environment. Furthermore $v_{e,z}, \dot{v}_{e,z}, \dot{e}_{xy}, e_{xy} \in \mathcal{L}_1 \cap \mathcal{L}_\infty$ while $f_{e,o}, \tau_e \in \mathcal{L}_\infty$.

*Proof:* The analysis for the x- and y- axes remains the same as in Section IV-B but the model of the environmental force has the damping term due to nonzero $B_e$ as shown in (12). The time derivative of the desired dynamics of (23) in the z-axis can be written as follows

$$m_{d,z}\ddot{v}_{e,z} = -b_{dz}\dot{v}_{e,z} + z_v^T \dot{R}_o^e f_{e,o} + z_v^T R_o^e \dot{f}_{e,o} \qquad (63)$$

Given that $\dot{f}_{e,o}$ can be written as

$$\dot{f}_{e,o} = -K_e R_e^o v_e + B_e(R_e^o \dot{v}_e + R_e^o[\omega_e]_\times v_e)$$

the second order dynamics can of (63) can be compactly written after algebraic manipulations as follows

$$m_{d,z}\ddot{v}_{e,z} = -b_{dz}v_{e,z} - k_e v_{e,z} + G_1 + G_2$$

where $G_1$ has been previously defined while $G_2$ is an additional term defined and bounded, respectively, in (79) and (80). It is now clear to see from (46) and (80) that $G_1 + G_2$ can be norm bounded as follows

$$\|G_1 + G_2\| \le \max(\gamma_1, \gamma_2, \gamma_3, \gamma_4) \|x\|$$

where $\gamma_4, \gamma_5, \gamma_6$ are positive constants. It is now possible to following the analysis as pursued in Section IV-B for the non-dissipative case to obtain exponential stability in the sense that $\|x\|, \|v_{e,z}\|, \|\dot{v}_{e,z}\| \le \rho e^{-\rho_1 t}$ where $\rho$ and $\rho_2$ are positive constants of analysis. Since $v_e, v_o, \dot{v}_{e,z} \in \mathcal{L}_\infty$, it can be seen from (42) and (12) that $\dot{z}_v^T R_o^e \dot{z}_v k_e(z_n - z_0) \in \mathcal{L}_\infty$ from which it is clear that $k_e(z_n - z_0) \in \mathcal{L}_\infty$ since $\dot{z}_v^T R_o^e \hat{z}_v > 0$ as shown in the Appendix. This implies from (12) and (13) that $f_{e,o}, f_{e,e}, \tau_e \in \mathcal{L}_\infty$. It can now be seen from (23) that $\lim_{t \to \infty} f_{e,z} = f_{dz}$. ∎

*Remark 8:* To summarize the analysis for the dissipative case, Lemma 1 shows that we can project the desired impedance on the frictional environment which is then followed up by Lemmas 6 and 7 which show that the robot end-effector aligns with the environment and tracks the end-effector velocity along the environmental tangential axes while also applying the desired amount of force in the environmental normal direction.

Given the above Lemmas, the passivity result for the proposed controller can be stated in the following theorem:

*Theorem 9:* The proposed control law can ensure the work done by the robot over and above the dissipation expected from relative motion at the desired surface velocity between the end-effector and the frictional environment(human) is limited in the sense that $W = \int_0^\infty \left( (F_{env})^T v_e - v_d^T B_e v_d \right) dt \le c < \infty$ where $B_e v_{d,xy}^T v_{d,xy} \ge 0$ is the essential power expended.

*Proof:* As similarly done in the proof of Theorem 5, we split the work done by the robot into two phases: (**Part a**) Work done before reaching the desired dynamics: Proof is identical as for the frictionless environment shown earlier. (**Part b**) After reaching the desired dynamics: The work done by the robot on the frictional environment is denoted by $W_b(t)$, and can be bounded as follows

$$
\begin{aligned}
W_b &= \int_{t_1}^\infty f_{e,xy}^T v_{e,xy} dt + \int_{t_1}^\infty f_{e,z}^T v_{e,z} dt + \int_{t_1}^\infty \tau_e^T \omega_e dt \\
&= \int_{t_1}^\infty f_{e,xy}^T \left( e_{v,xy} + v_{d,xy} \right) dt + \int_{t_1}^\infty f_{e,z}^T v_{e,z} dt \\
&\quad + \int_{t_1}^\infty \tau_e^T \omega_e dt \qquad (64) \\
&\le \int_{t_1}^\infty f_{e,xy}^T v_{d,xy} dt + \sup_t \{\|f_{e,xy}\|\} \int_{t_1}^\infty \|e_{v,xy}\| \, dt \\
&\quad + \sup_t \{\|f_{e,z}\|\} \int_{t_1}^\infty \|v_{e,z}\| \, dt \qquad (65) \\
&\quad + \sup_t \{\|\tau_e\|\} \int_{t_1}^\infty \|\omega_e\| \, dt \qquad (66)
\end{aligned}
$$

since $f_{e,xy}, f_{e,z}, \tau_e \in \mathcal{L}_\infty$ as previously shown. After some algebraic manipulations as shown in the Appendix, we can bound the first time in the above inequality as follows

$$
\begin{aligned}
\int_{t_1}^\infty f_{e,xy}^T v_{d,xy} dt &\le \int_{t_1}^\infty v_d^T B_e v_d dt + b_e \|v_{d,xy}\| \int_{t_1}^\infty \|e_{v,xy}\| \, dt \\
&\quad + \|v_{d,xy}\| \|A_{xy}\| \sup \left\{ \left( \|B_e\| \|v_e\| + \|f_{e,o}\|^2 \right) \right\} \int_{t_1}^\infty \|H_2\| \, dt \qquad (67)
\end{aligned}
$$

Since $H_2, e_{v,xy}, v_{e,z}, \omega_e \in \mathcal{L}_1$, we can clearly see from (65) and (67) that $W_b + \int_{t_1}^\infty v_d^T B_e v_d dt \le c < \infty$. Putting together the results from Part a and Part b proves the passivity result as stated in the statement of the theorem. ∎

### C. Simulation Results

In order to closely mirror experimental reality, we add joint friction, measurement noise, imperfect gravity compensation, and imperfect robot inertial matrix to the simulation studies. The joint friction model utilized is as follows

$$
\begin{aligned}
\tau_f &= F_c sng(q_i)[1 - \exp(\frac{-\dot{q}_i^2}{v_s})] \qquad (68) \\
&\quad + F_{si} sng(\dot{q}_i) \exp(\frac{-\dot{q}_i^2}{v_s}) + F_v \dot{q}_i
\end{aligned}
$$

where $F_c, F_s, F_v$ are the Coulomb, static, and viscous friction coefficients while $v_s$ is the Stribeck parameter. We also add measurement noise which follows a normal distribution with $\mu = -0.0001, \sigma = 0.0315$. We assume there is 3% imperfect gravity compensation in the simulation. As for the imperfect robot inertial matrix, we assume there is a constant 20% error

| | | | | | | |
|---|---|---|---|---|---|---|
| $F_c = diag$ [ 0.07 | 0.07 | 0.07 | 0.07 | 0.014 | 0.014 | 0.0035 ] |
| $F_s = diag$ [ 0.14 | 0.14 | 0.14 | 0.14 | 0.028 | 0.028 | 0.007 ] |
| $F_v = diag$ [ 0.13 | 0.13 | 0.13 | 0.13 | 0.026 | 0.026 | 0.013 ] |
| $v_s = diag$ [ 0.01 | 0.01 | 0.01 | 0.01 | 0.01 | 0.005 | 0.005 ] |
| $M_{d,v} = diag$ [ 1 | 1 | 10 ] | | | | |
| $I_d = diag$ [ 0.3 | 0.3 | 0.3 ] | | | | |
| $B_{d,v} = diag$ [ 10 | 10 | 70 ] | | | | |
| $K_d = diag$ [ 30 | 30 | 0 | 0 | 0 | 0 ] | |
| $Q = diag$ [ 24 | 48 | 48 | 60 | 72 | 72 | 84 ] |
| $P = diag$ [ 4 | 4 | 4 ] | | | | |
| $K_1 = diag$ [ 10 | 10 | 10 ] | | | | |
| $K_e = diag$ [ 0 | 0 | 506 ] | | | | |
| $B_e = diag$ [ 100 | 100 | 0 ] | | | | |

in the inertial matrix for the last three joints. We assume a planar surface for the simulation. We also replace the $sign(S)$ function in the sliding mode control with a continues function $\tanh(S)$ to mitigate the chattering phenomenon. The parameters utilized for each joint are listed in Table I.

During implementation, we treat $\theta_c$ and $\theta_d$ defined in (55) and (56) as immeasurable since is impractical to assume that the lever arm $r_e$ (as shown in Figure 2) is measurable. Instead, we replace the unknown value of $\|r_e\|$ with a known upperbound $\|r_m\|$ and redefine $\mathbf{q}(\acute{q}, \acute{q}_0)$ and $\mathbf{q}_d(\acute{q}_d, \acute{q}_{d0})$ as follows

$$
\acute{q}_0 = \cos(\frac{\acute{\theta}_c}{2}), \acute{q} = \sin(\frac{\acute{\theta}_c}{2}) n_e \qquad (69)
$$

$$
\acute{q}_{d0} = \cos(\frac{\acute{\theta}_d}{2}), \acute{q}_d = \sin(\frac{\acute{\theta}_d}{2}) n_d \qquad (70)
$$

where $n_e$ and $n_d$ have been previously defined, while $\sin(\acute{\theta}_c) \triangleq k_1 \sin(\theta_c) = \frac{\|r_e\|}{\|r_M\|\|f_{e,c}\|}$, $k_1 \triangleq \frac{\|r_e\|}{\|r_M\|}$, $\sin(\acute{\theta}_d) \triangleq k_2 \sin(\theta_d) = \frac{\|r_d\|}{\|r_M\|\|f_{e,e}\|}$, $k_2 \triangleq \frac{\|r_d\|}{\|r_M\|}$. Here, $k_2 \le 1$ is a constant, while $k_2 \le k_1 \le 1$ is a function of $\theta_c$. While the proof using the redefined quaternions is beyond the scope of this manuscript, it can be easily shown (see Appendix) that exponential convergence of $\acute{\theta}_c$ to $\acute{\theta}_d$ implies the exponential convergence of $\theta_c$ to $\theta_d$.

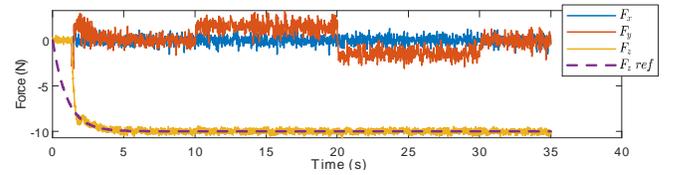

Fig. 10. Force tracking profile for frictional environment.

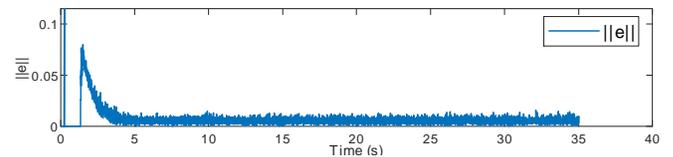

Fig. 11. Quaternion error tracking profile for frictional environment.

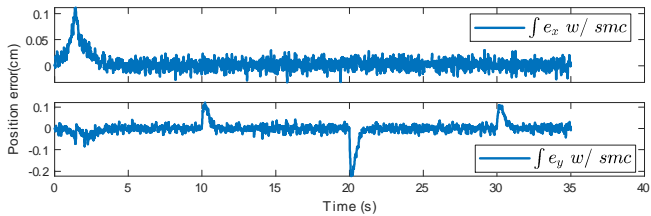

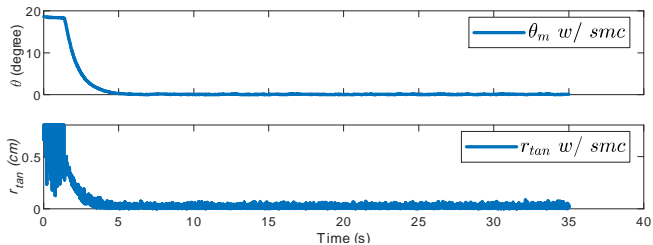

Fig. 12. Velocity tracking profile for frictional environment.

Fig. 13. Alignment evaluation for frictional environment.

From Figures 10-14, we can see similar performance as obtained with the frictionless environment. The force error in z-axis is regulated within 0.5N. The force in y-axis is the friction force from the environment surface. The quaternion error is also regulated to within 0.01 which means that the relative attitude between the end-effector z-axis and environment normal converges to ($e = 0, e_0 = 1$). Alignment results from Figure 13 show that the misalignment angle converges to 0. The velocity and position tracking in the x- and y-axes also perform as expected and the tracking error is regulated to within 0.2 cm. From Figure 14, we can see that the sliding mode control signal compensates for the disturbance, thereby converging the system dynamics to the desired dynamics.

## VI. EXPERIMENT

### A. Implementation

For experimentation, the Baxter robot from Rethink Robotics was used as the testbed. The ATI Mini45 force/torque sensor was mounted on the wrist of the Baxter to sense the interaction force/torque. The sensor was covered with soft rubber to lower the stiffness of the end-tool. As for the signal processing, we utilized an averaging filter with 45-sample window on the 1000 Hz Baxter status publish node. For ease of implementation, we modified the controllers (16) and (61) as follows

$$a_j = J^+ a_x - KS - Q \tanh(S) \quad (71)$$
$$a_{x,\omega} = RI_d^{-1}[-B_{d\omega}\omega_e + e_\tau] \quad (72)$$

where $e_\tau \triangleq \tau_e - \tau_d, \tau_d \triangleq r_d \times f_{e,e}, K$ is a constant diagonal matrix, while $I_d \triangleq diag\{i_d, i_d, i_d\} > 0$, and $B_{d\omega} \triangleq diag\{b_{d,\omega x}, b_{d,\omega y}, b_{d,\omega z}\} > 0$ denote desired rotational inertia and damping matrices.

### B. Experimental Results

The following experiments were conducted: 1) alignment of the end-effector with a yoga ball, 2) moving the end-effector

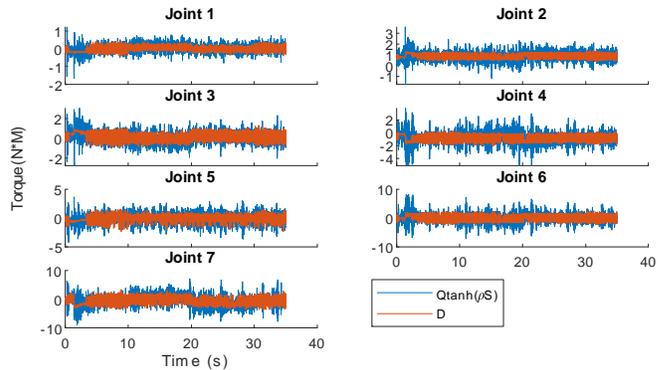

Fig. 14. SMC performance for frictional environment.

along the yoga ball, and 3) moving the end-effector along the back of a mannequin.

The first experiment focused on alignment and stabilization of the end-effector with respect to the environment in order to highlight the critical role of sliding-mode control (SMC). Here, we commanded 0 cm/s desired x- and y- velocities to the end-effector and $10(1 - e^{-0.2t})$ N as the desired force in the end-effector z-direction. From Figures 15 - 19, it can be seen that use of SMC leads to the interaction force along the end-effector z-direction being regulated to around 10 N with $\pm 0.5$ N error while the position error was less than 0.2 cm after the end-effector aligned with the surface. The figures also show that the system does not converge without application of SMC. The interaction torque was seen to decrease to less than 0.005 N · m, and the equivalent lever converged to around 0.3 mm; both results clearly illustrate that the end-effector aligned along the yoga ball surface with high fidelity through the course of the experiment.

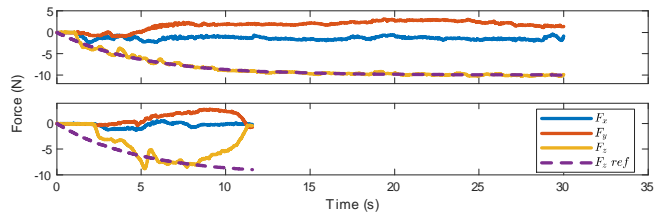

Fig. 15. External force for ball alignment.

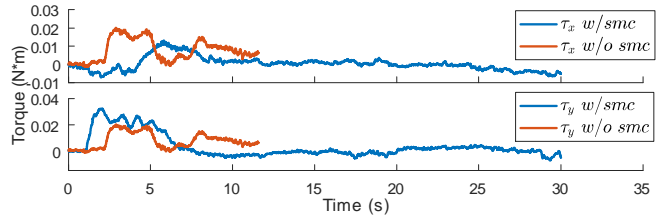

Fig. 16. External torque for ball alignment.

In the second experiment, we commanded $10(1 - e^{-0.2t})$ N as the desired force in the end-effector z-direction and the desired velocity was set to 0 from 0-10s for the initial alignment and $\pm 1.5\,\mathrm{cm/s}$ along the end-effector x-direction during the movement. The movement of the manipulator is shown in Figure 20.

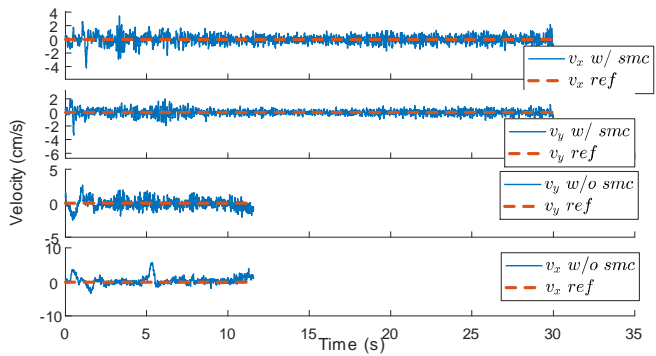

Fig. 17.   Velocity tracking for ball environment alignment experiment.

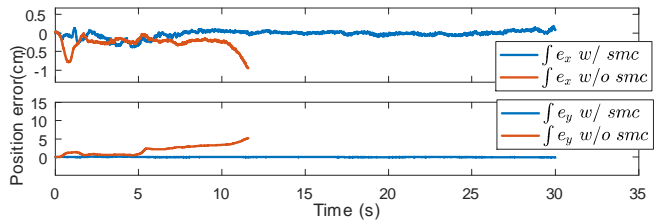

Fig. 18.   Position tracking and error for ball environment alignment experiment.

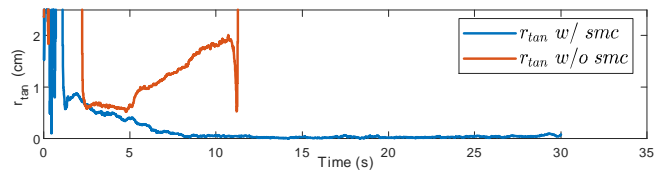

Fig. 19.   Alignment evaluation for ball environment alignment.

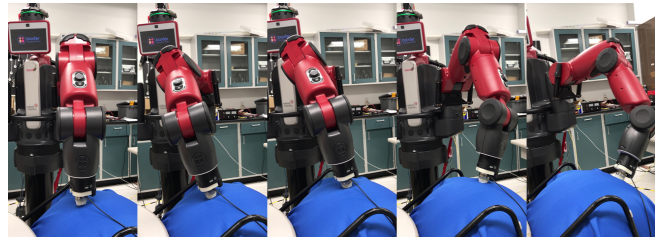

Fig. 20.   Video frames from the experiment on the yoga ball environment.

From Figures 21-25, the force along the end-effector z-direction was seen to be regulated around 10N while moving on the yoga ball with the desired velocity/position. It can be noticed that there was a minuscule torque error on the pitch axis of around $\pm 0.015\,\mathrm{N \cdot m}$ during movement. Since the controller design assumes a level environment, this torque error drives the end-effector to perform a constant angular velocity to align with the curvature of surface during the movement.

For the third experiment, we replaced the ball environment with a mannequin with an irregular surface typical of the back of a human torso. The movement of the manipulator is shown in Figure 26. From 0-10 s, we commanded the same force along the end-effector z-direction. Then, we set $v_{dx} = 0\,\mathrm{cm/s}$, $v_{dy} = 0.15\,\mathrm{cm/s}$ for 10-30 s, $v_{dx} = 0.05\,\mathrm{cm/s}$, $v_{dy} = 0\,\mathrm{cm/s}$ for 30-50 s, $v_{dx} = 0\,\mathrm{cm/s}$, $v_{dy} = 0\,\mathrm{cm/s}$ for 50-55 s, and then commanded the same velocities in the opposite direction. From Figures 27-31, it can be seen that the overall position/velocity tracking and force regulation performance was close to expected. However, the performance was comparatively degraded with respect to the results in the second experiment due to the high stiffness and the sharp curvature changes in the surface of the mannequin. During the time periods when the end-effector moves over the sharp curvature regions on the mannequin back, such as 13 s-18 s, 40 s-45 s, 70 s-73 s, 88 s-93 s, the force tracking and position tracking and alignment were affected. The force in end-effector z-direction reduced to 4N for about 0.1s, the position error increased to 0.6cm, the equivalent lever also increased to between 1 and 1.5 cm, but the degradation was seen to happen for short periods and the force and position/velocity tracking profiles recover each time.

## VII. Conclusions

In this paper, a robust hybrid controller has been implemented relying on the wrist force/torque and robot joint position/velocity feedback. A Lyapunov-based stability analysis is provided to prove both convergence as well as passivity of the interaction to ensure both performance and safety. Simulation as well as experimental results verify the performance and robustness of the proposed impedance controller in the presence of dynamic uncertainties as well as safety compliance of physical human-robot interactions for a redundant robot manipulator.

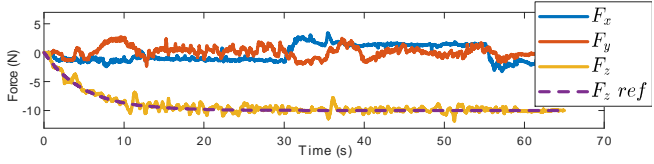

Fig. 21. External force for moving on ball environment.

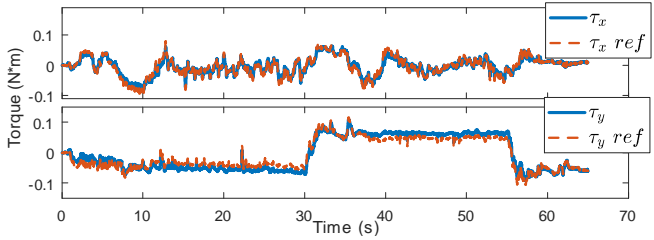

Fig. 22. External torque profile for moving on ball environment.

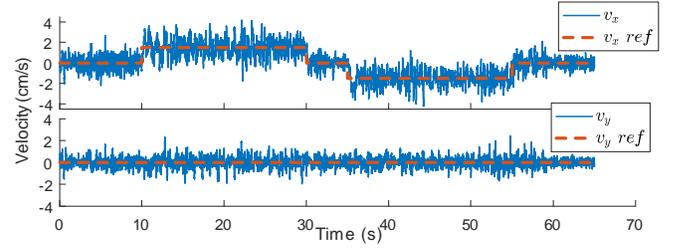

Fig. 23. Velocity tracking for moving on ball environment.

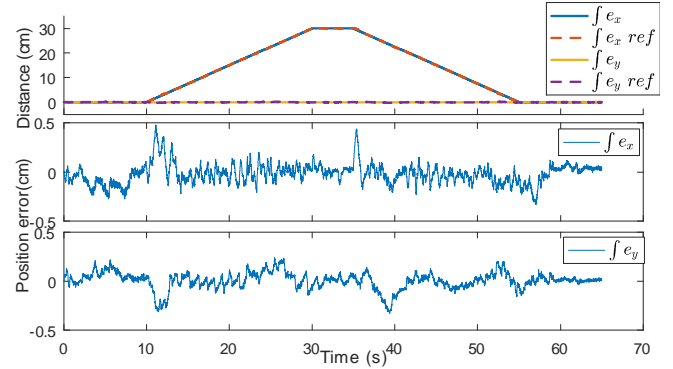

Fig. 24. Position tracking for moving on ball environment.

# Definition and Growth Bound of $G_1(t)$

The time derivative of (42) can be written as

$$
\begin{aligned}
m_{d,z}\ddot{v}_{e,z} &= -b_{dz}\dot{v}_{e,z} + \hat{z}_v^T \dot{R}_o^e f_{e,o} + \hat{z}_v^T R_o^e \dot{f}_{e,o} \\
&= -b_{dz}\dot{v}_{e,z} - \hat{z}_v^T [\omega_e]_\times R_o^e \hat{z}_v k_e(z_n - z_o) \\
&\quad - \hat{z}_v^T R_o^e K_e R_e^o v_e \\
&= -b_{dz}\dot{v}_{e,z} - \hat{z}_v^T R_o^e K_e R_e^o v_e \\
&\quad - \hat{z}_v^T [\omega_e]_\times R_o^e \hat{z}_v \frac{m_{d,z}\dot{v}_{e,z} + b_{dz}v_{e,z} + f_{d,z}}{\hat{z}_v^T R_o^e \hat{z}_v}
\end{aligned}
$$

where we have utilized (42) and (43) to substitute for $k_e(z_n - z_o)$ as well as the identity that $\dot{R}_o^e = -[\omega_e]_\times R_o^e$. By utilizing the identity that $R_o^e = I + H$, we can simplify

$$
m_{d,z}\ddot{v}_{e,z} = -b_{dz}\dot{v}_{e,z} - k_e v_{e,z} + G_1
$$

where

$$
\begin{aligned}
G_1 = &-k_e \hat{z}_v^T H^T v_e - k_e \hat{z}_v^T H \hat{z}_v v_{e,z} - k_e \hat{z}_v^T H \hat{z}_v \hat{z}_v^T H^T v_e \\
&-\hat{z}_v^T [\omega_e]_\times R_o^e \hat{z}_v \frac{m_{d,z}\dot{v}_{e,z} + b_{dz}v_{e,z} + f_{d,z}}{\hat{z}_v^T R_o^e \hat{z}_v}
\end{aligned} \tag{73}
$$

and

$$
H = -2q^T q I + 2qq^T - 2q_0[q]_\times
$$

which can be bounded as

$$
\|H\| \leq 4\|q\|^2 + 2\|q_0\|\|q\| \leq 6\|q\|
$$

which is an exponential bound owing to the exponential boundedness for $q$ from Lemma 2. By decomposing $v_e$ as

$$
v_e = \hat{z}_v v_{e,z} + \underbrace{\begin{bmatrix} v_x & v_y & 0 \end{bmatrix}^T}_{v_{e,xy0}},
$$

# Appendix

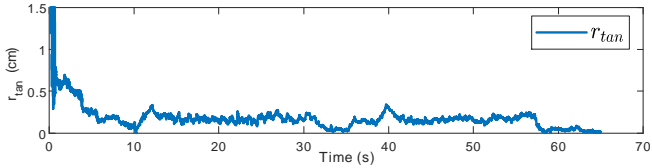

Fig. 25. Alignment evaluation for moving on ball environment.

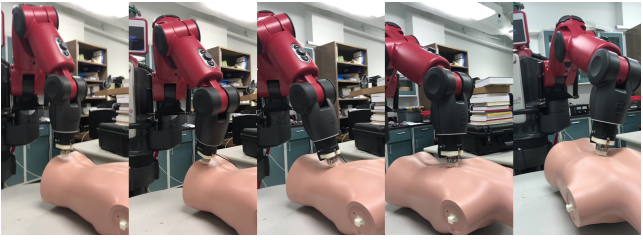

Fig. 26. Video frames from the experiment on the mannequin environment.

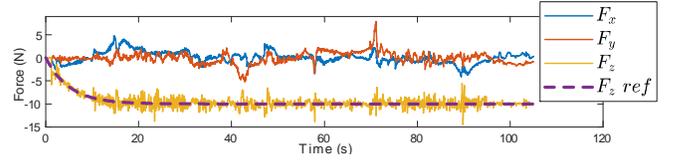

Fig. 27. External force profile for moving on mannequin back. During initial alignment (0-10s), force error is less than 0.5N. During movement, force error stays less than 2N except when end-effector moves on regions of sharp curvature where force error goes to 4N for less than 0.2s.

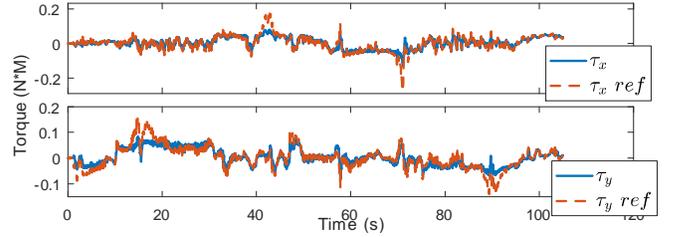

Fig. 28. External torque profile for moving on mannequin. To orient end-effector during changes in surface curvature, torque error increases up to 0.1N*m, otherwise remaining regulated to within 0.05N*m.

we can rewrite the perturbation system (73) as follows

$$G_1 = G_{11} + G_{12} + G_{13}$$

where

$$G_{11} = -\hat{z}_v^T [\omega_e]_\times R_o^e \hat{z}_v \frac{m_{d,z}\dot{v}_{e,z}}{\hat{z}_v^T R_o^e \hat{z}_v}$$

$$\begin{aligned} G_{12} = & -k_e \hat{z}_v^T H^T \hat{z}_v v_{e,z} - k_e \hat{z}_v^T H A v_{e,z} \\ & -k_e \hat{z}_v^T H \hat{z}_v \hat{z}_v^T H^T \hat{z}_v v_{e,z} - \hat{z}_v^T [\omega_e]_\times R_o^e \hat{z}_v \frac{b_{dz} v_{e,z}}{\hat{z}_v^T R_o^e \hat{z}_v} \end{aligned}$$

$$\begin{aligned} G_{13} = & -k_e \hat{z}_v^T H \hat{z}_v \hat{z}_v^T H^T v_{e,xy0} - k_e \hat{z}_v^T H^T v_{e,xy0} \\ & -\hat{z}_v^T [\omega_e]_\times R_o^e \hat{z}_v \frac{f_{d,z}}{\hat{z}_v^T R_o^e \hat{z}_v}. \end{aligned}$$

The denominator term $\hat{z}_v^T R_o^e \hat{z}_v$ in $G_{11}, G_{12}$, and $G_{13}$ can be lowerbounded as follows

$$\begin{aligned} \hat{z}_v^T R_o^e \hat{z}_v &= 2q_0^2 - 1 + 2q_3^2 \\ &\geq 2q_0^2(0) - 1 > 0 \end{aligned}$$

since $\|q_0(0)\| \in \left(1/\sqrt{2}, 1\right]$ and $q_0$ converges exponentially to 1. Since $\omega_e$ and $H$ converge exponentially to the origin, we can upperbound $G_{11}, G_{12}, G_{13}$ as follows

$$\|G_{11}\| \leq \gamma_1(m_{d,z}) \exp(-\gamma t) \|\dot{v}_{e,z}\| \tag{74}$$

$$\|G_{12}\| \leq \gamma_2(k_e, b_{dz}) \exp(-\gamma t) \|v_{e,z}\| \tag{75}$$

$$\|G_{13}\| \leq \gamma_3(k_e) \exp(-\gamma t) \tag{76}$$

where $\gamma_1(m_{d,z}), \gamma_2(k_e, b_{dz}), \gamma_3(k_e)$ are system parameter dependent constants.

## Definitions of $A$, $P$ and $Q$

The matrices $A$, $P$, and $Q$ are defined as follows

$$A \triangleq A = \begin{bmatrix} -\frac{b_{dz}}{m_{d,z}} & -\frac{k_e}{m_{d,z}} & 0 \\ 1 & 0 & 0 \\ 0 & 0 & -\gamma \end{bmatrix}$$

$$P \triangleq P = \begin{bmatrix} m_{d,z} & \epsilon m_{d,z} & 0 \\ \epsilon m_{d,z} & k_e & 0 \\ 0 & 0 & 1 \end{bmatrix} > 0 \tag{77}$$

$$Q \triangleq Q = \begin{bmatrix} b_{dz} - \epsilon m_{d,z} & \frac{1}{2}\epsilon b_{dz} & 0 \\ \frac{1}{2}\epsilon b_{dz} & \epsilon k_e & 0 \\ 0 & 0 & \gamma \end{bmatrix} > 0 \tag{78}$$

for sufficiently small $\epsilon$.

## Definition and Linear Growth Bound of $G_2(t)$

$$\begin{aligned} G_2 = & -\frac{\hat{z}_v^T [\omega_e]_\times R_o^e \hat{z}_v \hat{z}_v^T R_o^e B_e R_e^o v_e}{\hat{z}_v^T R_o^e \hat{z}_v} - \hat{z}_v^T [\omega_e]_\times R_o^e B_e R_e^o v_e \\ & +\hat{z}_v^T R_o^e B_e R_e^o [\omega_e]_\times v_e + \hat{z}_v^T H B_e \dot{v}_e + \hat{z}_v^T H B_e H^T \dot{v}_e \\ & +\hat{z}_v^T B_e H^T \dot{v}_e \end{aligned} \tag{79}$$

Since $\omega_e$ and $H$ converge exponentially to the origin and $v_x, v_y \in \mathcal{L}_\infty$, $G_2$ can be norm bounded (in a manner akin to $G_1$) as follows

$$\|G_2\| \leq \gamma_4(B_e) \exp(-\gamma t) \|x\| \tag{80}$$

where $\gamma_4(B_e)$ is a system parameter dependent constant.

## Bound for $\int_{t_1}^{\infty} -f_{e,xy}^T v_{d,xy} dt$

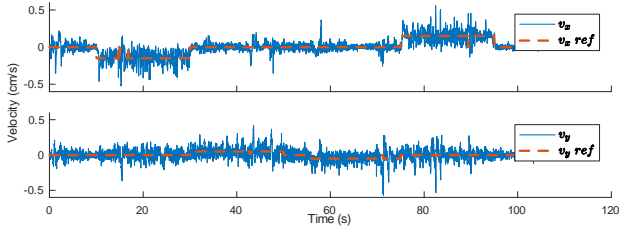

Fig. 29. Velocity tracking for moving on mannequin back.

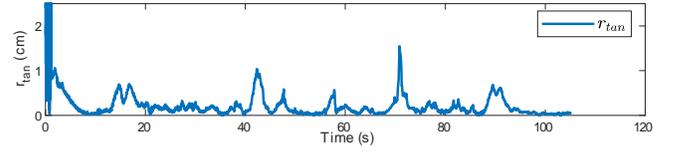

Fig. 31. Equivalent lever in the tangential plane of end-effector. At 13s, 41s, 70s and 88s, end-effector moves on regions of sharp curvature on mannequin which affects alignment during the transient, but the controller always recovers alignment in steady-state.

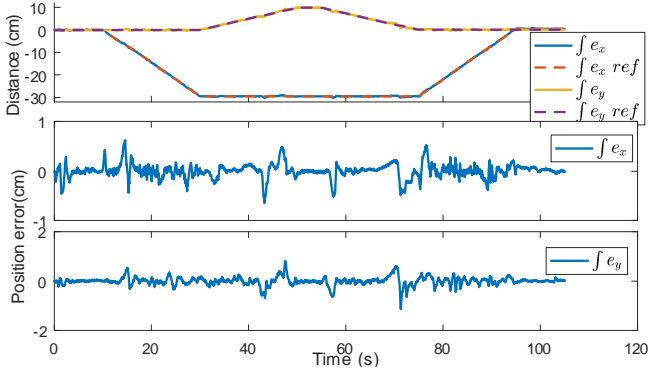

Fig. 30. Position tracking and error profile for moving on mannequin back.

## Exponential Convergence of $\acute{\theta}_c$ to $\acute{\theta}_d$ implies exponential convergence of $\theta_c$ to $\theta_d$

Based on the relation between $\acute{\theta}_c, \acute{\theta}_d$ and $\theta_c, \theta_d$ we can write

$$
\begin{aligned}
e &= q_{d0}q - q_0 q_d + [q]_\times q_d \\
&= \cos(\frac{\theta_d}{2})\sin(\frac{\theta_c}{2})n - \cos(\frac{\theta_c}{2})\sin(\frac{\theta_d}{2})n_d \\
&\quad + \sin(\frac{\theta_c}{2})\sin(\frac{\theta_d}{2})n \times n_d \\
&= \left[\frac{\cos(\frac{\theta_d}{2})\cos(\frac{\acute{\theta}_c}{2})}{k_1 \cos(\frac{\acute{\theta}_c}{2})\cos(\frac{\acute{\theta}_d}{2})}\right]\cos(\frac{\acute{\theta}_d}{2})\sin(\frac{\acute{\theta}_c}{2})n \\
&\quad - \left[\frac{\cos(\frac{\theta_c}{2})\cos(\frac{\acute{\theta}_d}{2})}{k_2 \cos(\frac{\acute{\theta}_c}{2})\cos(\frac{\theta_d}{2})}\right]\cos(\frac{\acute{\theta}_c}{2})\sin(\frac{\acute{\theta}_d}{2})n_d \\
&\quad + \left[\frac{\cos(\frac{\acute{\theta}_c}{2})\cos(\frac{\acute{\theta}_d}{2})}{k_1 k_2 \cos(\frac{\theta_c}{2})\cos(\frac{\theta_d}{2})}\right]\sin(\frac{\acute{\theta}_c}{2})\sin(\frac{\acute{\theta}_d}{2})n \times n_d \quad (81)
\end{aligned}
$$

Since $0 < k_2 \le k_1 \le 1$, and $\acute{\theta}_c, \acute{\theta}_d, \theta_c, \theta_d \in [0, \frac{\pi}{2})$, it is easy to see that all the bracketed terms in (81) can be bounded by constants $\lambda_1, \lambda_2, \lambda_3$ such that

$$
\begin{aligned}
\|e\| &\le \lambda_1 \cos(\frac{\acute{\theta}_d}{2})\sin(\frac{\acute{\theta}_c}{2}) + \lambda_2 \cos(\frac{\acute{\theta}_c}{2})\sin(\frac{\acute{\theta}_d}{2}) \\
&\quad + \lambda_3 \sin(\frac{\acute{\theta}_c}{2})\sin(\frac{\acute{\theta}_d}{2}) \\
&\le \max\{\lambda_1, \lambda_2, \lambda_3\}\|\acute{e}\|
\end{aligned}
$$

which clearly shows that exponential convergence of $\acute{e}$ to the origin implies the same convergence result for $e$.

$$
\begin{aligned}
&\int_{t_1}^{\infty} -f_{e,xy}^T v_{d,xy} dt \\
&= \int_{t_1}^{\infty} -(A_{xy}(I + H_2)f_{e,o})^T v_{d,xy} dt \\
&= \int_{t_1}^{\infty} -v_e^T (I + H_2)B_e v_{d,xy} dt - \int_{t_1}^{\infty} f_{e,o}^T H_2^T A_{xy}^T f_{e,o} v_{d,xy} dt \\
&= \int_{t_1}^{\infty} -(v_d + v_e)^T B_e A_{xy}^T v_{d,xy} dt - \int_{t_1}^{\infty} v_e^T H_2 B_e A_{xy}^T v_{d,xy} dt \\
&\quad - \int_{t_1}^{\infty} f_{e,o}^T H_2^T A_{xy}^T f_{e,o} v_{d,xy} dt \\
&= \int_{t_1}^{\infty} -v_d^T B_e v_d dt - \int_{t_1}^{\infty} e_v^T B_e A_{xy}^T v_{d,xy} dt \\
&\quad - \int_{t_1}^{\infty} v_e^T H_2 B_e A_{xy}^T v_{d,xy} dt - \int_{t_1}^{\infty} f_{e,o}^T H_2^T A_{xy}^T f_{e,o} v_{d,xy} dt \\
&= \int_{t_1}^{\infty} -v_d^T B_e v_d dt - \int_{t_1}^{\infty} b_e e_{v,xy}^T v_{d,xy} dt \\
&\quad - \int_{t_1}^{\infty} v_e^T H_2 B_e A_{xy}^T v_{d,xy} dt - \int_{t_1}^{\infty} f_{e,o}^T H_2^T A_{xy}^T f_{e,o} v_{d,xy} dt \\
&\le \int_{t_1}^{\infty} -v_d^T B_e v_d dt + b_e \|v_{d,xy}\| \int_{t_1}^{\infty} \|e_{v,xy}\| dt \\
&\quad + \|v_{d,xy}\|\|A_{xy}\| \sup\left\{\left(\|B_e\|\|v_e\| + \|f_{e,o}\|^2\right)\right\}\int_{t_1}^{\infty}\|H_2\| dt
\end{aligned}
$$

where we have utilized the fact that $R_o^e = \underbrace{I - 2e^T e I + 2ee^T - 2e_0[e]_\times}_{H_2}$.